\documentclass[letterpaper, 10 pt, conference]{ieeeconf}  

\IEEEoverridecommandlockouts                              

\usepackage{amsmath} 
\usepackage{amssymb}  
\usepackage{lettrine}
\usepackage{kotex}
\usepackage{graphicx}
\usepackage{algorithm}
\usepackage{algpseudocode}
\usepackage{multirow}
\usepackage{booktabs}
\usepackage{adjustbox}
\usepackage{subfigure}
\usepackage{cite}
\usepackage{hyperref}
\usepackage{bbold} 

\title{\LARGE \bf
GIN: Graph-based Interaction-aware Constraint Policy Optimization for Autonomous Driving
}

\author{Se-Wook Yoo, Chan Kim, Jin-Woo Choi, Seong-Woo Kim, and Seung-Woo Seo.
    \thanks{This work was supported by the Basic Science Research Program through the National Research Foundation of Korea (NRF) funded by the Ministry of Science ICT(2017R1E1A1A01075171).}
    \thanks{The authors are with Seoul National University, Seoul 08826, South Korea. S. Seo is the corresponding author. (E-mail:
        {\small tpdnr1360@snu.ac.kr, chan\_kim@snu.ac.kr, wlsdn9530@snu.ac.kr, snwoo@snu.ac.kr, sseo@snu.ac.kr})
    }%
}

\begin{document}

\maketitle

\begin{abstract}

Applying reinforcement learning to autonomous driving entails particular challenges, primarily due to dynamically changing traffic flows. To address such challenges, it is necessary to quickly determine response strategies to the changing intentions of surrounding vehicles. This paper proposes a new policy optimization method for safe driving using graph-based interaction-aware constraints. In this framework, the motion prediction and control modules are trained simultaneously while sharing a latent representation that contains a social context. To reflect social interactions, we illustrate the movements of agents in graph form and filter the features with the graph convolution networks. This helps preserve the spatiotemporal locality of adjacent nodes. Furthermore, we create feedback loops to combine these two modules effectively. As a result, this approach encourages the learned controller to be safe from dynamic risks and renders the motion prediction robust to abnormal movements. In the experiment, we set up a navigation scenario comprising various situations with CARLA, an urban driving simulator. The experiments show state-of-the-art performance on navigation strategy and motion prediction compared to the baselines.

\end{abstract}



\section{INTRODUCTION}

\lettrine{I}{n} safety-critical systems such as autonomous driving, the autonomous vehicle must avoid dynamic obstacles without collision while following the lane along the road network. Recently, deep reinforcement learning (DRL) \cite{sallab2017deep} have shown general and efficient driving strategies by maximizing the reward for following the lane while providing a negative penalty for a collision. Nevertheless, this does not guarantee that the safety constraints are observed. To overcome this issue, a method based on safe reinforcement learning (RL) \cite{achiam2017constrained} is proposed to approximate the expected long-term costs as constraints for policy optimization. Although this method tends to maintain minimum constraints for the collision risks over static objects, the problem becomes complicated for dynamic objects. This is because it is difficult to infer the randomness of the cost distribution by providing sparse cost signals only when collisions occur. In urban driving, it is necessary to prevent potential risks by generating cost signals while considering the predicted motions between dynamic agents.

\begin{figure}[t]
    \centering
    \includegraphics[width=1.0\columnwidth]{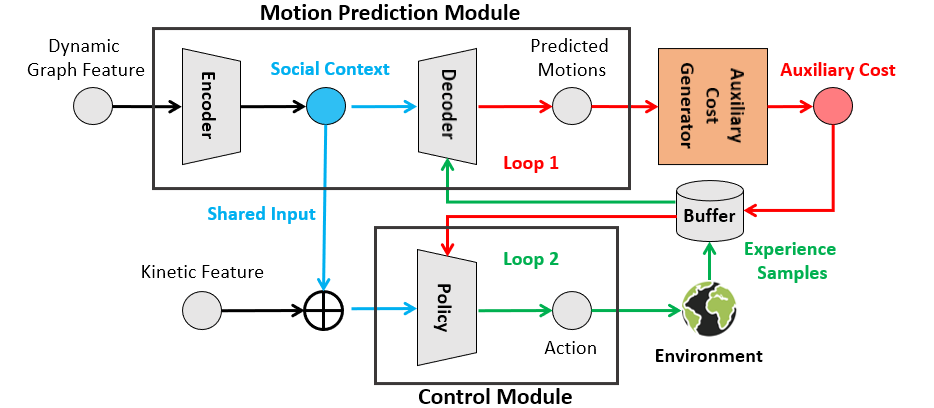}
    \caption{Proposed scheme. The blue line indicates the social context vector shared by the modules for motion prediction and control, while the combination of colored lines illustrates feedback loops between two modules. According to the red line, auxiliary costs are additional constraint signals for control. Moreover, the green line illustrates that the saved samples serve as ground truth for motion prediction.}
    \label{fig:overall}
    \vspace{-0.5cm}
\end{figure}

However, it is particularly difficult to predict motions in congested situations because the intentions of dynamic objects are not directly observable. To solve this problem, previous works \cite{houenou2013vehicle, yoon2020road} used expensive high-definition (HD) maps as prior knowledge to predict various motion patterns over different types of roads, such as multi-lane roads, multi-way intersections, or roundabouts. Nevertheless, These methods are not generally applied to changes in specific road context information such as traffic signs and road marks. Moreover, the accuracy of abnormal patterns decreases in case some agents do not conform with traffic rules. To address this issue, \cite{schreier2014bayesian} attempts to predict trajectories according to perceived maneuvers, such as lane changing, braking, or turning. Unfortunately, incorrect classification of maneuver types also leads to inaccurate predictions. To show robust prediction, \cite{deo2018convolutional, alahi2016social, li2019grip} reflect the interactions of social movements by learning latent variables that encode sequential information over objects. Although \cite{deo2018convolutional} shows high accuracy through social pooling, it incurs a high computational cost for multiple objects. Unlike this, \cite{li2019grip} expresses dynamic agents as graph form, and the spatial importance is considered in a graph convolution network (GCN) \cite{kipf2016semi}. This shows faster and more accurate results for multiple objects. However, prediction through GCN shows non-universal results for new graph structures that have not been seen in pre-collected data. Although we use GCN structure to understand the interaction for motion prediction, arbitrary graphs are dealt with using the data being collected by a learning RL agent.

In this paper, we present a new framework that understands the interaction of surrounding vehicles and makes appropriate decisions according to the detected risk. Fig. \ref{fig:overall} shows the overview of our scheme, which combines the modules for motion prediction and control to learn tactical driving strategies. Specifically, the two modules are simultaneously trained through the RL framework while sharing a social context vector that explains the interactions between agents. To learn the context vector, we spatiotemporally refine the dynamic graph features using an encoder. Using this context vector, we create two feedback loops to combine the two modules effectively. The first one generates auxiliary costs by geometrically interpreting the predicted motions, which are used as constraints for policy optimization. The other one utilizes samples collected by the policy during training as ground truth for the prediction. 

The main contributions of the proposed method are as follows:

\begin{itemize}
    \item We propose a new safe RL framework with a policy constraining interaction-aware risks by geometrically interpreting the predicted trajectories.
    \item We represent the movements of vehicles as a dynamic graph and train it to learn social interaction directly without prior knowledge. 
    \item We integrate the path prediction and control modules within a single RL framework by allowing the two modules to share spatiotemporally compressed social context for robust prediction and response to dynamic risks.
\end{itemize}

The remainder of this paper is organized as follows. Related works are reviewed in Section \uppercase\expandafter{\romannumeral2}. The background is explained in Section \uppercase\expandafter{\romannumeral3}. The proposed scheme and implementation details are presented in Section \uppercase\expandafter{\romannumeral4}. Our experimental results are presented in Section \uppercase\expandafter{\romannumeral5}. Finally, we conclude the paper in Section \uppercase\expandafter{\romannumeral6}.

\section{Related Work}
Our framework aims to learn a policy with safety constraints and predict motions while considering interactions. In this section, we discuss two related approaches, namely, safe RL and interaction-aware motion planning.

\subsection{Safe Reinforcement Learning}
In RL literature, the concept of safety is unavoidable in the safety-critical domain. Generally, the constraint Markov decision process (CMDP) \cite{altman1999constrained} is a natural approach for finding a solution among set of allowable policies that guarantee safety. \cite{achiam2017constrained} approximates the average cumulative costs under the CMDP using a policy gradient (PG) algorithm based on DRL. To adjust the trade-off caused by the different scales between the reward and cost, \cite{tessler2018reward} proposed an alternative penalty in which the designed cost function was multiplied by the learned adaptive weight. However, these methods do not account for the potential risk of the constraint threshold being exceeded in a heavy-tailed cost distribution. To address this problem, \cite{chow2017risk} replaces average cumulative costs with conditional value-at-risk (CVaR) \cite{rockafellar2000optimization}. This minimizes the heavy-tailed risk by predicting the full distribution. \cite{tang2019worst} derives the PG formula using the CVaR measure to show behaviors according to the risk level. Nevertheless, it has only been applied to a few scenarios static obstacles exist for a short period.

In contrast, we focus on navigation problems with complex interactions between dynamic agents over long periods. Furthermore, we adopt a worst-case soft actor-critic (WCSAC) \cite{yang2021wcsac} method that interprets the CVaR criterion in terms of alternative penalties. This reduces the approximated error over the cost distribution by balancing the adaptive weights for entropy and safety. However, this method has several limitations. For example, if a conservative risk level is used, it becomes overly pessimistic. Moreover, if the cost signal is sparse, the variance of the distribution decreases. In this case, the CVaR criterion is close to the mean value. This still has limitations, especially when rare events pose a great risk. To overcome the problems, we create dense cost signals by combining the interaction-aware motion prediction process with the WCSAC policy optimization method.

\subsection{Interaction-aware Motion Planning}

Recently, several studies \cite{werling2010optimal, nilsson2015automated, damerow2015risk} proposed integrating RL and planning to learn driving skills. These works predict the trajectory of surrounding vehicles and then plan the trajectory of the ego-vehicle to show reactive behaviors but do not reflect the interaction during planning. Some methods predict model approximated by a partially observable Markov decision process (POMDP) \cite{kochenderfer2015decision} to obtain a tactical driving strategy. For example, \cite{hoel2019combining} predicts a temporally correlated region using a search tree created through a Monte Carlo tree search (MCTS). Meanwhile, \cite{kim2021stfp} receives temporal sequences of occupancy grid maps (OGMs) for implicit intention reasoning and \cite{wang2021reinforcement} predicts the time horizon for motion planning. Besides, there have been approaches to encourage cooperative and competitive behaviors in a multi-agent system. \cite{qi2018intent} approximates the sum of the influence of other agents and their intent as a linear function. \cite{jiang2018graph} uses a GCN-based Q-network that shares the observations of all agents. However, it is costly to use sophisticated inputs such as OGM or shared observations. On the contrary, interaction-aware prediction methods do not require expensive resources. Although some methods using social pooling \cite{deo2018convolutional, alahi2016social} have shown high accuracy, these are practically intractable for predicting multiple objects. Meanwhile, in \cite{gao2020vectornet}, road contexts, such as lane and dynamic agents, were expressed as graphs for intuitive interpretation. In this method, after filtering graph features through GCN, the learned features capture high-dimensional global graph relationships that maintain semantic and spatial locality. Similarly, in \cite{li2019grip}, the surrounding agents are expressed in a graph form. This makes the prediction for multiple objects accurate and rapid. Hence, we adopt the structure of this method to reflect interactions that maintain spatiotemporal locality.

Unlike previous studies, in which interaction was reflected only in motion prediction, we consider the interaction to create constraints for control. To combine motion prediction and policy models efficiently, we learn spatiotemporal compressed representation, which both models share as input. In addition, we create a meaningful connection between the prediction model and policy through the auxiliary costs generated by the geometric analysis of the predicted motions. Eventually, we find an optimal policy constrained by the predicted interaction-aware long-term costs for safety.

\begin{figure*}[t]
    \vspace{0.5cm}
    \centering
    \includegraphics[width=0.95\linewidth]{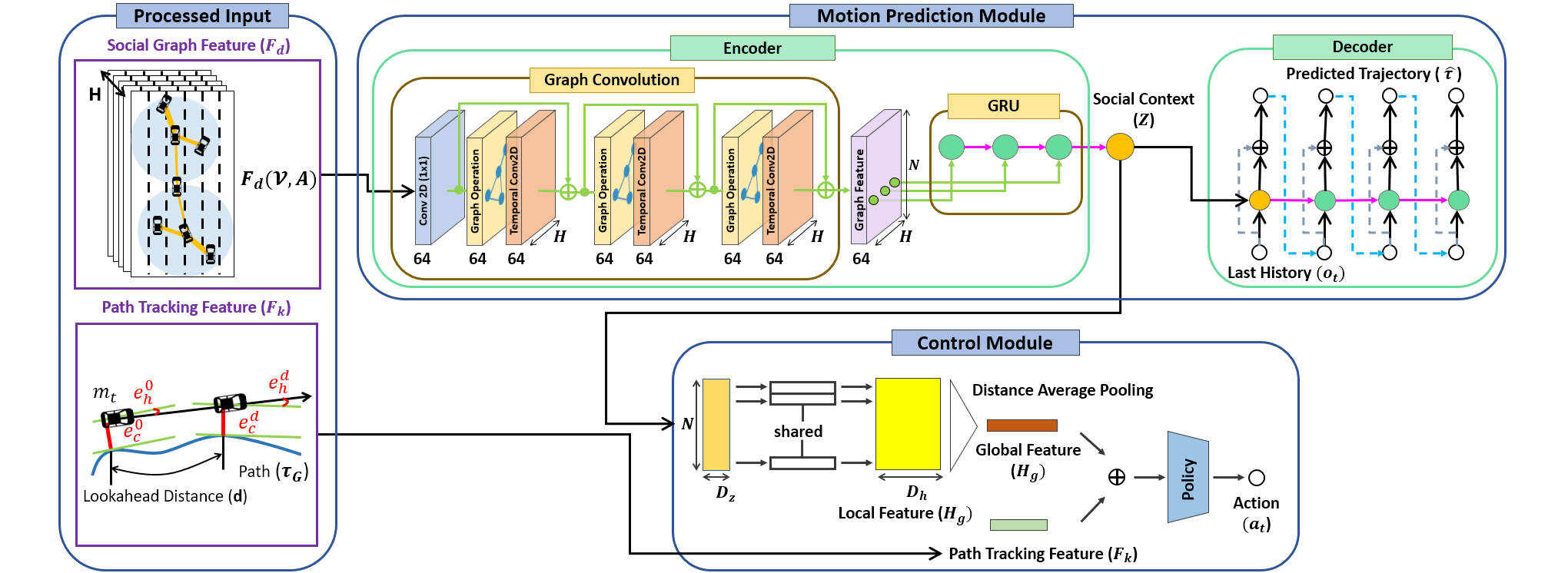}
    \caption{Overall architecture of proposed framework. Motion prediction is performed by variational autoencoding models while creating intermediate representation that explains the social context. The control module receives input features for tracking a path and understanding social context. Then, it outputs an action that considers social interaction.}
    \label{fig:architecture}
    \vspace{-0.5cm}
\end{figure*}

\section{Preliminaries}

\subsection{Constrained Markov Decision Processes (CMDP)}
The CMDP is an extension of the MDP, is utilized for safe RL. The function $\mathcal{C} : \mathcal{S} \times \mathcal{A} \rightarrow \mathbb{R}^{+}$ of costs $c$ is added to a tuple of MDP $(\mathcal{S}, \mathcal{A}, \mathcal{R}, \mathcal{T}, \gamma)$, where $\mathcal{S}$ is a set of states $s$, $\mathcal{A}$ is a set of actions $a$, $\mathcal{R}: \mathcal{S} \times \mathcal{A} \rightarrow \mathbb{R}$ is a function of rewards $r$, $T : \mathcal{S} \times \mathcal{A} \times \mathcal{S} \rightarrow [0,1]$ is a state transition probability distribution, $\gamma \in (0,1)$ is a discount factor, and $\pi:\mathcal{S} \times \mathcal{A} \rightarrow$ [0,1] is a policy distribution. The objective of safe RL is to find an optimal policy in a set of allowable policies that satisfy the constraints. This maximizes the expected return, such that the expected long-term cost remains below given threshold $d$, which is formulated in \cite{achiam2017constrained} as follows:

\begin{equation}
 \mathbb{E}_{\pi}[\sum_{t}^{\infty} \gamma^{t}r(s_t, a_t)] \; s.t.\; \mathbb{E}_{\pi}\;[\sum_{t}^{\infty} \gamma^{t}c(s_t, a_t)] \leq d
\end{equation}

\subsection{Worst-Case SAC (WCSAC)}

Limitation of maximizing the average value is that it does not consider the high potential risks caused by the heavy-tail of cost distribution. To reduce the heavy-tail risk, the WCSAC algorithm approximates the mean $Q_{\pi}^{c}(s,a)$ and variance $V_{\pi}^{c}(s,a)$ of the cost distribution $C_{\pi}(s,a)$ as a Gaussian distribution $\mathcal{N}(Q_{\pi}^{c}(s,a), V_{\pi}^{c}(s,a))$, and then replaces the existing measure with a closed-form estimation for CVaR. The risk-sensitive criterion $\Gamma_{\pi}$ for the risk level $\alpha \in (0,1)$ is calculated in \cite{tang2019worst} as follows:

\begin{equation}
    \Gamma_{\pi}^{\alpha} 
    \doteq \mbox{CVaR}_{\alpha} 
    = Q_{\pi}^{c} (s,a) + \alpha^{-1} \phi(\Phi(\alpha))\sqrt{V_{\pi}^{c}(s,a)},
\end{equation}

\noindent where $\phi(\cdot)$ and $\Phi(\cdot)$ are the PDF and CDF, respectively, of the standard normal distribution. The objective of WCSAC is to encourage safe exploration by auto-tuning adaptive entropy $\beta$ and safety weight $\kappa$ under the problem of maximum entropy RL \cite{haarnoja2017reinforcement} with safety constraints.
For all times $t$ and given risk level $\alpha$, the objective of the training policy is derived in \cite{yang2021wcsac} as follows:

\begin{equation}
    \mathbb{E}_{\pi} [\beta \log \pi(a_t \vert s_t) - X_{\alpha,\kappa}^{\pi}(s_t,a_t)]
    \; s.t. \; \Gamma_{\pi}^{\alpha}(s_t,a_t) \leq d,
    \label{eq:wcsac_objective}
\end{equation}

\noindent  where $X_{\alpha, \kappa}^{\pi} (s,a) = Q_{\pi}^{r}(s,a) - \kappa \Gamma_{\pi}^{\alpha}(s,a)$.

\subsection{Graph Convolution Network (GCN)}

As a variant of a convolutional neural network (CNN) \cite{krizhevsky2012imagenet}, GCN reflects flexible adjacency to handle abstract concepts such as social interaction. This method extracts valuable information from an undirected graph $\mathcal{G} = (\mathcal{V,E})$. The node features $v_{i} \in \mathcal{V} \in \mathbb{R}^{N \times D_{v}}$ are filtered by aggregating information from adjacent nodes along the edges $e_{ij} = (v_{i},v_{j}) \in \mathcal{E}$ using the following propagation rule \cite{kipf2016semi}:

\begin{equation}
    \psi(A,H^{l};W^{l})
    = \sigma (\tilde{D}^{-\frac{1}{2}}\tilde{A}\tilde{D}^{-\frac{1}{2}}H^{l}W^{l}),
    \label{eq:graph_operation}
\end{equation}

\noindent where an adjacency matrix with added self-connections is denoted by $\tilde{A} = A + I_{N} \in \mathbb{R}^{N \times N}$, the degree matrix of $\tilde{A}$ is $\tilde{D}_{ii} =\sum_{j}\tilde{A}_{ij}$, $W^{l}$ is a layer-specific trainable weight, $\sigma(\cdot)$ denotes an activation function, and $H^{l} \in \mathbb{R}^{N \times D_{h}}$ is the embedding vector in the $l^{th}$ layer except for $H^{0}=X$.

\section{Method}
We introduce a framework that predicts interaction-aware motion and controls an ego-vehicle to respond to dynamically changing driving styles of the surrounding vehicles. Fig. 2 shows the detailed structure of each part. This section describes how these three modules are used: input preprocessing, interaction-aware motion prediction, and interaction-aware control.

\subsection{Input Preprocessing}

We use two input features to perform navigation to the destination. One is a path tracking feature $F_{k}$. This is required to track the given global path. The second is the social graph feature $F_{d}$, which accounts for the dynamic environment. 

\subsubsection{Path Tracking Feature}

The path tracking feature $F_{k} \in \mathbb{R}^{5}$ is created using both online and offline information. Specifically, localization information $m_{t}$ is received through the GPS in real-time, and road network information $\mathcal{G}_{n}$ is obtained from a database query within a specific area. A global path $\tau_{G}$ that follows $\mathcal{G}_{n}$ is generated in advance from the current location to the target location using the A* algorithm \cite{keirsey1984algorithm}. Subsequently, the cross-tracking $e_{c}$ and heading angle $e_{h}$ errors are calculated at every timestamp for the current location and the location before the look-ahead distance $d$ utilizing $\tau_{G}$ and $m_{t}$. Finally, the current speed of the ego-vehicle is added. 

\begin{figure}
    \vspace{0.2cm}
    \centering
    \includegraphics[width=0.8\columnwidth]{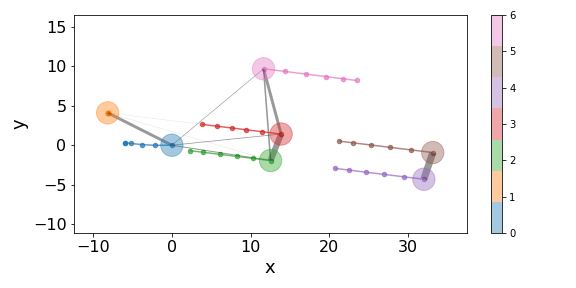}
    \vspace{-0.4cm}
    \caption{Visualization of the social graph feature. The value of each color indicates the index of surrounding vehicles. The inter-frame edges are described with the same color as that for each vehicle and depict the progress of movement. The spatial edges expressed using gray colors indicate proximity, where the width indicates the size of the weight.}
    \label{fig:designed_graph}
    \vspace{-0.5cm}
\end{figure}

\subsubsection{Social Graph Feature}

We represent the movement of dynamic agents on the road in graph form $F_{d}$, which consists of a node $\mathcal{V}$ and an adjacency matrix $\mathbf{A}$. $\mathbf{A}$ is constructed using a manually designed rule, as shown in Fig. \ref{fig:designed_graph}. The node feature is denoted by $\mathcal{V} \in \mathbb{R}^{N \times H \times D_{v}} =\{v_{it} \vert \mbox{ for }i = 1,\dots, N \mbox{, and } t = 1, \dots, H\}$, where N is the number of surrounding vehicles, and H is the history length. Each node vector comprises the position and heading angle in the egocentric coordinate system, relative velocity, relative acceleration, relative angular velocity, and mask. A mask indicates whether an object exists within the recognition range. The edges comprise inter-frame $E_{F}=\{(v_{it}, v_{i(t+1)}) \vert \mbox{ for } i=1,\dots,N \mbox{, and } t=1, \dots, H-1 \}$ and spatial edges $E_{S}=\{e_{ij} = (v_{iH}, v_{jH}) \vert\ \mbox{if } d_{ij} \leq D_{close} \mbox{ for } i,j = 1,\dots,N \}$, where $d_{ij}$ is the Euclidean distance between two vertices in the last time frame. $D_{close}$ is the distance threshold for considering neighbors. The adjacency of the edges is denoted as $A \in \mathbb{R}^{N \times N}$. The elements $a_{ij}$ of $A$ are obtained using the indicator function $\mathbb{1}_{\{\exists e_{ij} \in E_{S} \} }$. To learn the high-order relation of the interactions, we use $\mathbf{A} \in \mathbb{R}^{L \times N \times N} = \{ A_{k} \vert \; k = 1, \dots, L \}$ with multi-hop, where $L$ is the maximum number of hops. The $k^{th}$ element of $\mathbf{A}$ is expressed as follows:

\begin{equation}
A_{k} = \{
A_{k}^{ij} 
= a_{kij}e^{-\frac{d_{ij}}{\tau}}
\vert\; i,j = 1, \dots N
\},
\label{eq:edge}
\end{equation}

\noindent where an element of $A$ to the $k^{th}$ power is denoted by $a_{kij}$, and the weights of the edges are modeled as a Boltzmann function with the temperature constant $\tau$.

\subsection{Interaction-aware Motion Prediction}

Trajectories are predicted by considering the interactions between ego-vehicle and other vehicles. We utilize $F_{d}$ as the input because it concisely represents dynamic agents instead of costly information such as OGM or HD maps. The structure of the graph changes over time as the agents enter or leave the recognition range. To enable the prediction module to reason an arbitrary graph inductively, the module is trained with sub-graph samples collected by the RL agent. The module for motion prediction consists of an encoder $\nu$ and decoder $\rho$, parameterized by $\eta$ and $\zeta$, respectively.

\subsubsection{Encoder}

In Fig. \ref{fig:architecture}, sub-graphs sampled from the replay buffer are used as inputs for the encoder $\nu_{\eta}$. For the sub-graphs, a bottleneck layer, which is a convolutional layer with a 1 by 1 kernel, is applied to increase the size of the node features. Subsequently, the features pass through three layers of the graph convolutional block to extract high-order relations that encourage sophisticated driving strategies. Each graph convolutional block compacts the latent representation while maintaining spatiotemporal connectivity. In this procedure, the sum of the graph operation $\sum_{k=1}^{L} \psi(A_{k},H^{l} ; W_{k}^{l})$ in Eq. \ref{eq:graph_operation} extracts spatially correlated features along the spatial edges $E_{S}$. Next, a convolution operation is applied along the inter-frame edges $E_{F}$ to understand the temporal dependencies between the dynamic agents, which helps the learned intentions to remain consistent. Finally, a residual connection is utilized to prevent the gradient from vanishing. After multilayer blocks, to create a social context $Z \in \mathbb{R}^{N \times D_{z}} = (z_{1}, \dots, z_{N})$, the sequential features are summarized by the GRU model. 

\subsubsection{Decoder}
Based on the implied context $Z$, the decoder $\rho_{\zeta}$ generates the future trajectory $\hat{\tau}_{t}$ of the surrounding vehicles and allows the control module to know long-term risks. This is because it can provide signals for detected risks by geometrically interpreting the predicted trajectories. The initial hidden state of the decoder GRU is the social context, and the initial input is the current location $o_{t}$ of each vehicle. Each GRU cell predicts the velocity profiles of vehicle through a skip connection between the input and the output. The trajectories $\tau_{t}$ sampled from the replay buffer are used as the ground truth. The prediction module is trained using regression loss with p-norm, which is as follows:

\begin{equation}
    \mathcal{L}_{P} = \mathbb{E}_{\pi}[\frac{1}{H} \sum_{t=1}^{H}\vert\vert \rho_{\zeta}(\nu_{\eta}(F_{d}), o_{t}) - \tau_{t}\vert\vert_{p}],
    \label{eq:prediction_loss}
\end{equation}

\subsection{Interaction-aware Control}
Our control module outputs actions by considering properties for path tracking and collision avoidance in dynamic environment via the RL framework. For this reason, our method uses the path tracking feature $F_{k}$ and social context $Z$, represents the interaction of objects. To compress the social context in the view of the control of an ego-vehicle, we design a pooling module. Moreover, auxiliary costs are generated using the predicted motions. Finally, we explain how to optimize the modules for motion prediction and control simultaneously through the off-policy PG method under the CMDP.

\subsubsection{Distance Average Pooling}

Because the order of the nodes changes when creating a graph for dynamic objects within the recognition range, we conserve the permutational invariance of the social context $Z$. For elements of $Z$, the local features $H_{l} = (h_{1}, \dots, h_{N}) \in \mathbb{R}^{N \times D_{h}}$ are obtained through a feed-forward network that shares weights. Then, distance average pooling (DAP) is applied to create a global feature $H_{g} \in \mathbb{R}^{D_{h}}$. In terms of the ego-vehicle, the closer the agents are, the more the intentions should be reflected. Therefore, we design the DAP as follows:

\begin{equation}
    H_{g} = \frac{1}{N}\sum_{i=1}^{N} h_{i} \odot e^{\frac{-\sqrt{x_{i}^{2} + y_{i}^{2}}}{\tau}},
\label{eq:dap}
\end{equation}

\noindent where $\tau$ regulates the influence according to the distance, $(x_{i},y_{i})$ is the current location of the $i^{th}$ vehicle in the egocentric coordinate system, and $\odot$ is defined as the Hadamard product operation is performed by broadcasting each distance weight to $h_{i}$. After DAP, we concatenate the global feature $H_{g}$ and hand-designed path tracking feature ${F_{k}}$, and pass $[H_{g},F_{k}]$ as input to the policy $\pi$, parameterized by $\theta$. 

\begin{figure}[!t]
    \vspace{0.4cm}
    \centering
    \subfigure[Polygon-Polygons]{\includegraphics[width=0.40\columnwidth]{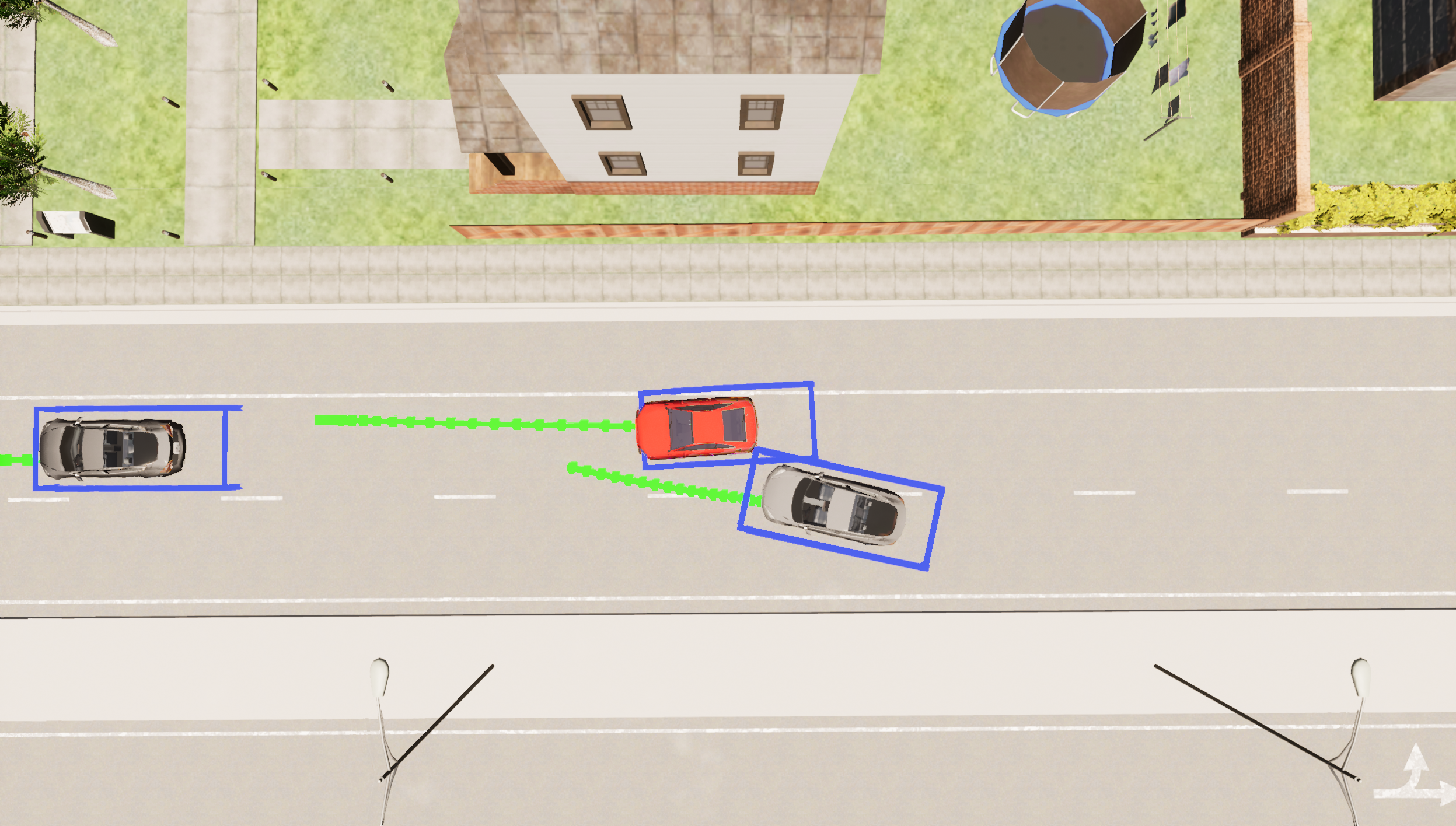}}
    \subfigure[Polygon-Polylines]{\includegraphics[width=0.40\columnwidth]{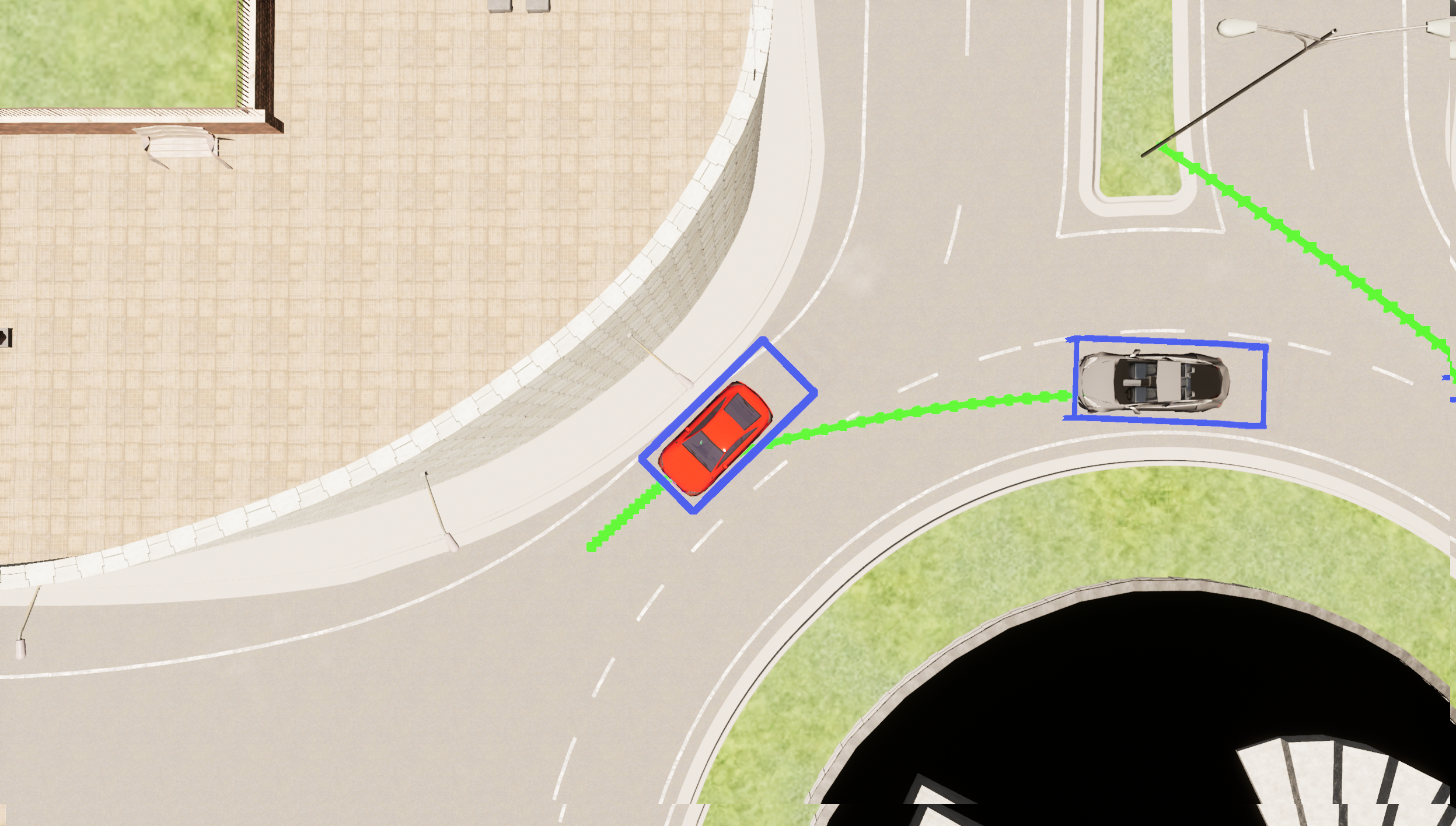}}
    \subfigure[Polyline-Polygons]{\includegraphics[width=0.40\columnwidth]{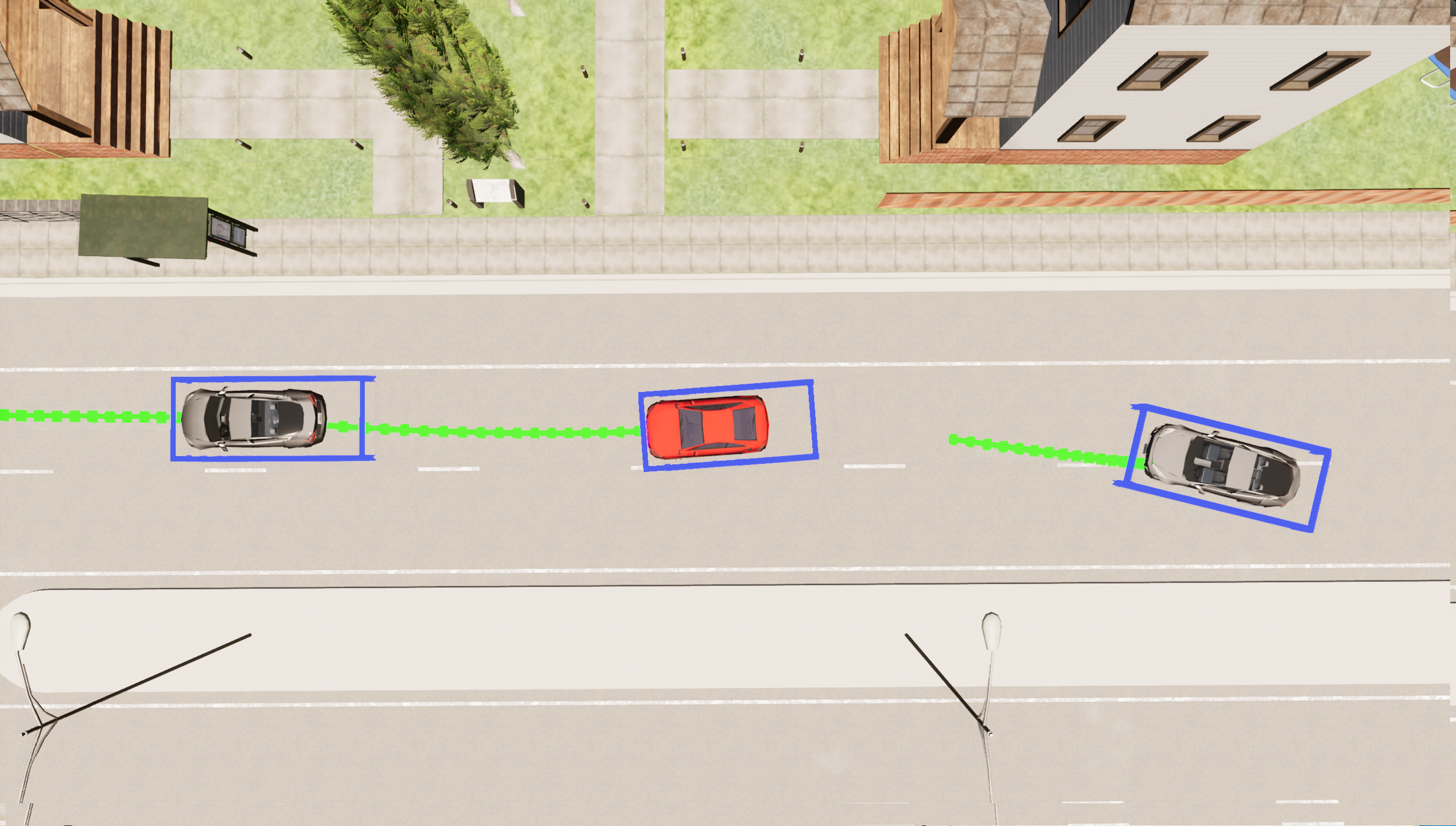}}
    \subfigure[Polyline-Polylines]{\includegraphics[width=0.40\columnwidth]{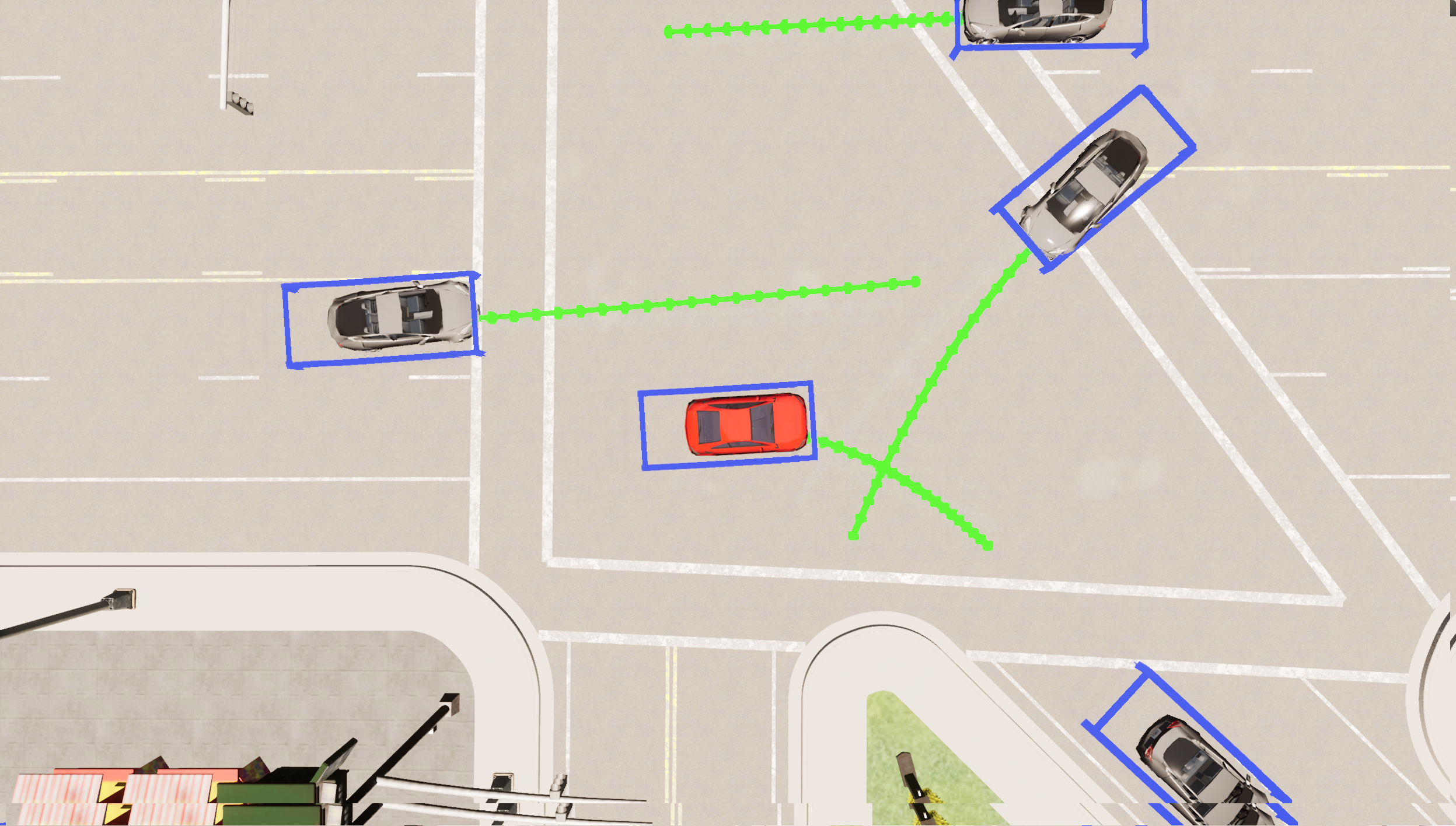}}
    \caption{Four cases of social risk. The ego-vehicle is red, and the others are gray. Boundaries with safety margins are expressed by blue bounding boxes. The green polylines represent the predicted future trajectories.}
    \label{fig:social_risk}
    \vspace{-0.5cm}
\end{figure}

\subsubsection{Auxiliary Cost}
To prevent costs from being obtained only in the case of collision are sparse, we create an auxiliary cost by using the predicted trajectories. Thus, we change the sparse signals to dense ones, which promotes the learned policy $\pi_{\theta}$ to avoid long-term risks. We divide the risk detection into four cases, as shown in Fig \ref{fig:social_risk}. To interpret the social risk geometrically, we used polygons and polylines. The polygon is a local bounding box with safety margin $(\epsilon_{l}, \epsilon_{w})$ added to the length and width of each vehicle. The polyline represents the trajectory of each vehicle as predicted by the decoder $\mathcal{\rho}_{\zeta}$. For the case Fig. \ref{fig:social_risk}(a),  we check the overlap between the polygon of the ego-vehicle and others using the separating axis theorem (SAT) \cite{gottschalk1996separating}. In the cases shown in Fig. \ref{fig:social_risk}(b-d), we utilize a method \cite{chazelle1992optimal} to check the intersection between two-line segments based on a counter-clockwise (CCW) algorithm. To create line segments, we connect the starting and ending points of the polygons. For the polygons, we use each side as a line segment. The auxiliary cost is defined as follows:

\begin{equation}
    \hat{c}_{\mathcal{\rho}} =
    \mathbb{1}_{\{\exists G \cap \mathbf{G}\}} \vee
    \mathbb{1}_{\{\exists G \cap \mathbf{L}\}} \vee 
    \mathbb{1}_{\{\exists L \cap \mathbf{G}\}} \vee 
    \mathbb{1}_{\{\exists L \cap \mathbf{L}\}}, 
    \label{eq:auxiliary_cost}
\end{equation}

\noindent  where $G$ and $L$ are the polygon and polyline, respectively, of the ego-vehicle. The bold fonts such as $\mathbf{G}$ and $\mathbf{L}$ are those of the other vehicles. $\cap$ represents a set of intersections between ego-ones and others, and $\vee$ is an OR operator. 

\begin{figure}[!t]
    \centering
    \begin{minipage}{\columnwidth}
        \begin{algorithm}[H]
            \caption{GIN}\label{alg:GIN}
            \label{algo}
            \begin{algorithmic}[1]
                \State \textbf{Initialize}: Parameters of networks $\pi_{\theta}, Q_{\psi}, C_{\mu}, \mathcal{\nu}_{\eta}, \mathcal{\rho}_{\zeta}$,
                
                
                \State \textbf{Initialize}: Adaptable weights $\beta, \kappa$
                
                \State \textbf{Initialize}: Replay buffer $\mathcal{D} \gets \emptyset$
                
                \State Copy target networks $<\bar{\theta}, \bar{\psi}, \bar{\mu}>$ with $<\theta, \psi, \mu>$
                
                \State Load road network graph $\mathcal{G}_{n}$ from database 
                
                \For{each iteration \do}
                    \State Randomly select $p_{s}$ and $p_{g}$
                    \State $\tau_{g} \gets \mbox{A}^{*}(p_{s},p_{g}, \mathcal{G}_{n})$
                    
                    \For{each environmental step \do}
                        \State $F_{d} \gets \phi_{d}(o_{t-H+1:t})$ and $F_{k} \gets \phi_{k}(\tau_{g}, m_{t})$
                        
                        \State $\Breve{F}_{d} \gets \phi_{d}(o_{t-2H+1:t})$ centered on $t-H$ time.
                        
                        \State $Z \gets \nu_{\eta}(F_{d})$
                        
                        \State $s_{t} = \{Z, F_{k}\}$
                        
                        \State $a_{t} \sim \pi_{\theta}(a_{t} \vert s_{t})$
                    
                        \State $s_{t+1} \sim \mathcal{T}(s_{t+1} \vert s_{t}, a_{t})$
                        
                        \State $\hat{\tau}_{t} \gets \rho_{\zeta}(Z, o_{t})$
                        
                        \State ${c}_{t}^{+} = c(s_{t},a_{t}) + \hat{c}(\hat{\tau}_{t})$
                        
                        \State $\mathcal{D} \gets \mathcal{D} \cap 
                        \{ s_{t}, a_{t}, r(s_{t},a_{t}), {c}_{t}^{+}, s_{t+1}, \Breve{F}_{d} \}$
                    
                    \EndFor
                    
                    \For{each gradient step \do}
                    
                        \State Randomly sample experience from $\mathcal{D}$
                        
                        \State Update $\eta, \zeta$ with $\nabla_{\eta, \zeta} \mathcal{L}_{p}(\Breve{F}_{d})$ from Eq. \ref{eq:prediction_loss}
                        
                        \State $\theta, \psi, \mu, \beta, \kappa$ $\gets$ WCSAC($s_{t}, a_{t}, r_{t}, {c}_{t}^{+}, s_{t+1}$)
                        
                        \State Update $<\bar{\theta}, \bar{\psi}, \bar{\mu}>$ with $<\theta, \psi, \mu>$
                    \EndFor
                \EndFor
            \end{algorithmic}
        \end{algorithm}
    \end{minipage}
    \vspace{-0.5cm}
\end{figure}

\begin{table*}[t]
    \vspace{0.4cm}
    \centering
    \caption{Comparison Average Results with standard deviation in Navigation Scenario }
    \vspace{-0.2cm}
    \begin{adjustbox}{max width=0.95\linewidth}
        \begin{tabular}{c c c c c c c c}
            \toprule
            \textsc{Method} & \textsc{Return} &  \textsc{Success Rate} & \textsc{Distance} & \textsc{Cost} & \textsc{Collision Rate} & \textsc{Survival Step} & \textsc{Success Step} \\ 
            \midrule
            SAC & 945 $\pm$ 598
                & 0.48 $\pm$ 0.07 
                & 122 $\pm$ 73
                & 11.5	$\pm$ 16.6
                & 0.44 $\pm$ 0.06 
                & 254 $\pm$ 148 
                & 366 $\pm$ 75 \\
            WCSAC & 984 $\pm$ 557 
                & 0.54 $\pm$ 0.11 
                & 130 $\pm$ 71
                & 7.2 $\pm $13.1
                & 0.35 $\pm$ 0.12 
                & \textbf{325} $\pm$ 179
                & 432 $\pm$ 86 \\
            GIN (Ours) & \textbf{1284 $\pm$ 488}
            (30.5\%$\uparrow$)
                & \textbf{0.76 $\pm$ 0.03}
                (40.7\%$\uparrow$) 
                & \textbf{161 $\pm$ 58}
                (23.8\%$\uparrow$) 
                & \textbf{4.0 $\pm$ 10.4}
                (44.4\%$\downarrow$) 
                & \textbf{0.19 $\pm$ 0.03} 
                (45.7\%$\downarrow$) 
                & 303 $\pm$ \textbf{119}
                (6.8\%$\downarrow$) 
                & \textbf{337 $\pm$ 70} 
                (7.9\%$\downarrow$) \\
            \bottomrule 
        \end{tabular}
    \end{adjustbox}
    \vspace{-0.5cm}
    \label{tab:navigation}
\end{table*}

\subsubsection{Optimization}

Here, the \textbf{G}raph-based \textbf{IN}tention-aware constraint policy optimization, namely \textbf{GIN}, is explained through Alg. \ref{algo}. Before training loop, we obtain a global path $\tau_{g}$ that follows road network $\mathcal{G}_{n}$ using the A* algorithm (lines 7, 8). At each step, the agent interacts with the environment, generates samples during training, and stores them in a buffer (lines 9-19). The input preprocessing modules are $\phi_{d}$ and $\phi_{k}$. They operate separately to create the social graph feature $F_{d}$ or the path tracking feature $F_{k}$ (line 10). $F_{d}$ describes dynamic objects through history of observations, and $F_{k}$ is to explain the kinematics of the system. Additionally, we accumulate another feature $\Breve{F}_{d}$ centered at the midpoint of length $2H$ in the time domain to use the data for predicting future trajectories $\hat{\tau}_{t}$ (line 11). Given the state processed by the encoder $\nu_{\eta}$, the agent executes an action, and proceeds to the next state (lines 12-15). To avoid long-term risks considering this interaction, we use the auxiliary cost $\hat{c_{\rho}}$ in Eq. \ref{eq:auxiliary_cost}. It is obtained using $\hat{\tau}_{t}$ obtained from decoder $\rho_{\zeta}$ (lines 16-18). For the gradient step, all model parameters are updated with batches sampled from $\mathcal{D}$ (lines 20-25). The encoder and decoder are updated using the prediction loss in Eq. \ref{eq:prediction_loss}. The parameters $<\theta, \psi, \mu>$ of the actor, critic, and approximated cost, and the adaptable weights $<\beta, \kappa>$ are learned by the WCSAC algorithm using accumulated samples through Eq. \ref{eq:wcsac_objective}. Finally, the parameters of the target networks $<\bar{\theta}, \bar{\psi}, \bar{\mu}>$ are updated with moving averages for learning stability.

\section{Experiments}

\begin{figure}[!t]
    \centering
    \subfigure[Town 03]{\includegraphics[width=0.38\columnwidth]{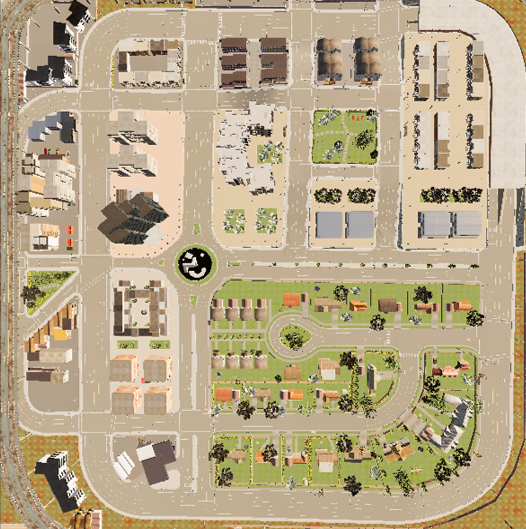}}
    \subfigure[Navigation]{\includegraphics[width=0.39\columnwidth]{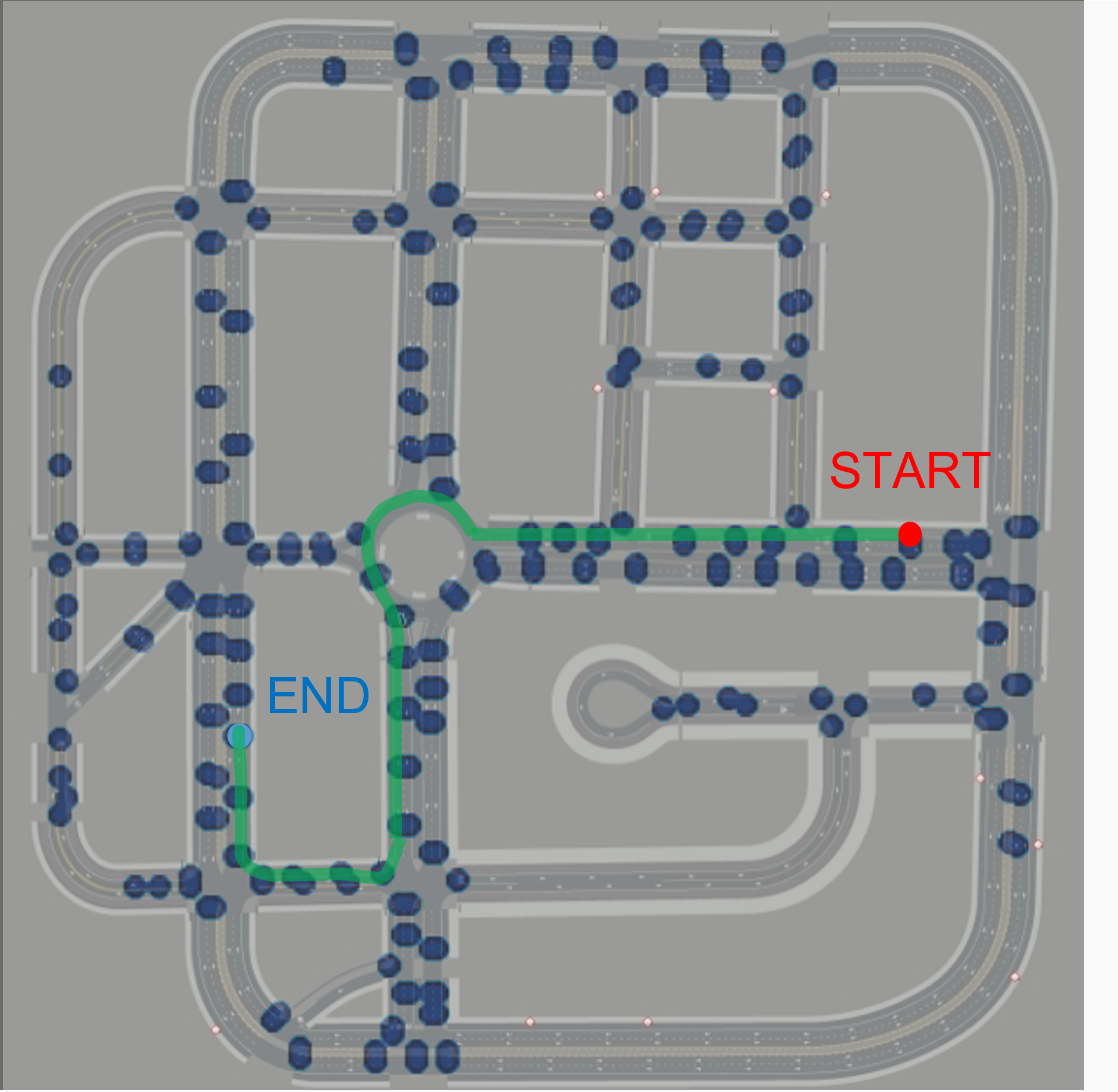}}

    \caption{Visualization of navigation environment: \textbf{(a)} is the map used in the virtual environment and \textbf{(b)} explains an episode. All possible spawn points are represented by navy dots. The start and end points are red and blue, respectively. The navigation route is indicated in green.}
    \label{fig:environment}
    \vspace{-0.5cm}
\end{figure}

\subsection{Simulation Settings}


To demonstrate the performance of the proposed method, we set up a navigation environment that contained diverse scenes utilizing the CARLA physics engine \cite{dosovitskiy2017carla}, as shown in Fig. \ref{fig:environment}. In our experiment, the following situations are assumed in the scenarios: a fixed route length is randomly selected, the speed difference percentage for each vehicle is 20\%, traffic lights and signs are ignored to create abnormal patterns. Finally, to track the global path obtained by the $A^{*}$ algorithm, the reward function $r$ is designed as follows:

\begin{equation}
    \resizebox{0.9\columnwidth}{!}{$
        r = 
        w_{1} \vert v_{s} \vert  
        + w_{2} \vert  v_{d} \vert
        + w_{3} \vert e_{c} \vert 
        + w_{4} \vert \Delta e_{c} \vert
        + w_{5} \vert \delta_{s} \vert
        + w_{6} \vert e_{h} \vert\vert v_{d} \vert,
    $}
    \label{eq:reward}    
\end{equation}

\noindent where $v_{s}$ and $v_{d}$ are the longitudinal and latitudinal velocities over the reference path, $e_{c}$ and $e_{h}$ are the cross-tracking and heading angle errors, respectively, and $\delta_{s}$ is the steering wheel angle. Finally, $(w_1,\dots, w_{6})$ are the experimentally obtained weights. To reflect the collision with the control, the environmental cost function is calculated using the normalized collision intensity. The code is available online\footnote{https://github.com/Usaywook/gin.git}.

\subsection{Metrics}

We set seven measurements to compare the extent to which the driving strategies are tactical. Return, distance, cost, and survival step are measured with the total cumulative rewards, traveled distance, cumulative costs, and steps, respectively, during an episode. Success rate is the number of episodes in which the destination is reached over all episodes. Collision rate is the number of episodes in which collisions occur over all episodes. Finally, the success step refers to the total number of steps taken over all successful episodes. Furthermore, to compare how accurately motions are predicted, we use the average displacement error (ADE) and final displacement error (FDE). All the above metrics are measured using the averages of five seeds over 100 episodes.

\subsection{Tactical Driving Strategy}

We evaluate performance in terms of efficiency and safety. The SAC and WCSAC algorithms based on the off-policy maximum entropy RL framework are adopted as baselines. For a fair comparison, baselines also use separate features for path tracking and dynamic objects as inputs. The difference is that they employ observations only in a single time frame. Subsequently, max pooling is applied to obtain the global features. Table. \ref{tab:navigation} shows the averages with standard deviations for all metrics for each method. WCSAC exhibits slightly better performance than SAC. This is because using the worst-case measurement reduces the probability of collision, as shown in the cost and collision rate. Nevertheless, a high percentage of crash situations still exists. This is because the long-term dynamic risks do not tend to be reflected due to sparse cost signals. Furthermore, although WCSAC increases the survival steps, the augmented values in the success step indicate a loss in terms of efficiency by driving conservatively.

\begin{figure}[t]
    \vspace{-0.3cm}
    \centering
    \subfigure[Success Rate]{\includegraphics[width=0.49\columnwidth]{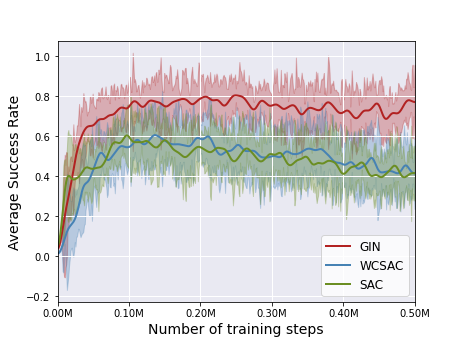}}
    \subfigure[Distance]{\includegraphics[width=0.49\columnwidth]{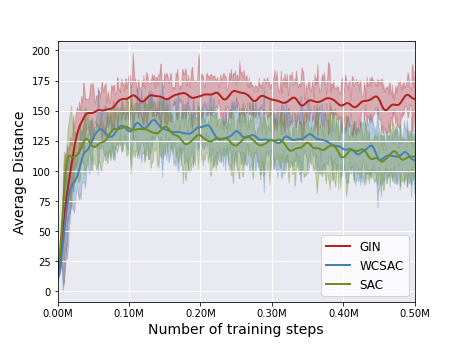}}
    \caption{Learning curve for navigation environment. The darker-colored lines and shaded areas denote the average return and standard deviations.}
    \label{fig:learning_curve}
    \vspace{-0.5cm}
\end{figure}

In contrast, our method shows dramatic improvements in all respects, compared to baselines. Although our method has a lower value than WCSAC in terms of survival steps, it has an efficient driving strategy in terms of a considerable decrease in reaching steps to the goal. Moreover, according to Fig. \ref{fig:learning_curve}, unlike baselines that show similar curves over success rate and collision rate, our method shows a significant improvement in sample efficiency and stability. This is because it considers social interactions by capturing the local connectivity between agents through the encoder. It also reduces the collision rate by utilizing the dense long-term cost signals predicted by decoders as additional constraints. Although SAC has a faster initial learning speed owing to more aggressive driving than WCSAC, conflicts become more frequent as learning progresses. Consequently, the performance gradually converges to a lower value. In contrast, WCSAC maintains a slightly better final performance than SAC.

\subsection{Robust Behavior Prediction}

To evaluate the performance of motion prediction, we adopt the vanilla-gated recurrent unit (V-GRU) and graph-based interaction-aware trajectory prediction (GRIP) algorithms as baselines. For GRIP, the control part is excluded from our method. In the case of V-GRU, the graph convolution part is excluded from the encoder in GRIP. V-GRU does not reflect regional interactions in the spatial domain. For both baselines, 200 successful episodes are used for training. Each episode is performed using the autopilot agent obtained from CARLA. To compare the generalizability of various scenes, 20 episodes that included both successes and failures are collected as test data.

According to Table. \ref{tab:prediction} GRIP reduces errors slightly in ADE compared to V-GRU but significantly reduces errors in FDE. This is because most vehicles interact longitudinally along lane. In contrast to GRIP, our method shows a greater improvement in both metrics. Notably, this performance is achieved even without the prepared training data. This is because it has the advantage of utilizing data that include various scenes obtained in the trial-and-error process of RL. In fact, it is difficult for existing graph-based methods to generalize the dynamic graph, whereas our method overcomes this difficulty by combining graph feature learning with RL.

\begin{table}[t]
    \vspace{0.4cm}
    \centering
    \caption{Comparison Trajectory Prediction Results}
    \vspace{-0.2cm}
    \begin{adjustbox}{max width=0.9\columnwidth}
    \begin{tabular}{c c c}
        \toprule
         \textsc{Method} & \textsc{ADE} & \textsc{FDE} \\
         \midrule
         V-GRU & 0.57 $\pm$ 0.10 & 1.45 $\pm$ 0.26 \\
         GRIP & 0.55 $\pm$ 0.07 & 1.32 $\pm$ 0.20 \\
         GIN (Ours) & \textbf{0.48 $\pm$ 0.06} 
         (12.7\%$\downarrow$)
         & \textbf{1.15 $\pm$ 0.19} 
         (12.9\%$\downarrow$)\\
         \bottomrule
    \end{tabular}
    \end{adjustbox}
    \label{tab:prediction}
    \vspace{-0.5cm}
\end{table}

\begin{table}[b]
    \vspace{-0.5cm}
    \centering
    \caption{Effect of interaction-aware planing}
    \vspace{-0.2cm}
    \begin{adjustbox}{max width=0.9\columnwidth}
        \begin{tabular}{c c c c}
            \toprule
            \textsc{Method} 
                & \textsc{Success Rate}
                & \textsc{Collision Rate}
                & \textsc{Success Step} \\
            \midrule
            RN & 0.37 $\pm$ 0.10
                & 0.31 $\pm$ 0.04 
                & 428.15 $\pm$ 91.47 \\
            HMM & 0.23 $\pm$ 0.07
                & 0.25 $\pm$ 0.11 
                & 508.78 $\pm$ 98.75 \\
            V-GRU & 0.19 $\pm$ 0.05
                & 0.26 $\pm$ 0.06 
                & 451.80 $\pm$ 93.04 \\
            SE & \textbf{0.74 $\pm$ 0.02}
                & \textbf{0.20 $\pm$ 0.05} 
                & \textbf{364.30 $\pm$ 77.29} \\
            \bottomrule
        \end{tabular}
    \end{adjustbox}
    \label{tab:interaction_aware_planing}
\end{table}

\subsection{Ablation Study}

To prove the effect of recognizing the topological characteristics of surrounding objects, we compare the performances while replacing our encoder that reflects social interaction, called social encoder (SE), with random noise (RN), a hidden Markov model (HMM), and the vanilla gated recurrent unit (V-GRU). For a fair comparison, Eq. \ref{eq:prediction_loss}, which trains the decoder, is excluded. Moreover, all policies are equivalently optimized using the WCSAC algorithm in Eq. \ref{eq:wcsac_objective}. Specifically, RN is sampled from the standard normal distribution. The HMM is pretrained with 500 samples of accident-free scenes. To apply the HMM, the observation variable is modeled as a Gaussian variable. Additionally, the social context is approximated using a categorical variable with 3-levels. Incidentally, V-GRU implies that the graph convolution module is excluded from the SE. As presented in Table. \ref{tab:interaction_aware_planing}, the SE shows a dramatic improvement over others. This is because the spatiotemporally abstracted features effectively express the dynamic interactions for tactical driving. In contrast, RN has better results than HMM and V-GRU. This is because the dynamic features are ignored and they focus on learning to follow the path. For V-GRU, the intent of distant objects interferes with the perception of the surrounding circumstances, which causes fatal consequences. In the case of HMM, it exhibits low performance owing to its conservative behavior. This is because it is difficult for HMM to adapt to an environment in which a collision exists.

\begin{table}[t]
    \vspace{0.4cm}
    \centering
    \caption{Changes in Performance while adjusting the model}
    \vspace{-0.2cm}
    \begin{adjustbox}{max width=0.95\linewidth}
        \begin{tabular}{c|c c c c|c}
            \toprule
            \textsc{Index} 
                & \textsc{Cost}
                & \textsc{Wedge} 
                & \textsc{DAP}
                & \textsc{Aux. Cost}
                & \textsc{Success Rate}  \\          
            \midrule
            B1 & N & N & N & N & 0.41 $\pm$ 0.03 \\
            B2 & N & \textbf{\underline{Y}} & N & N & 0.48 $\pm$ 0.08 \\
            B3 & N & Y & \textbf{\underline{Y}} & N & 0.71 $\pm$ 0.05 \\
            B4 & \textbf{\underline{Y}} & N & N & N & 0.50 $\pm$ 0.04 \\
            B5 & Y & \textbf{\underline{Y}} & N & N & 0.55 $\pm$ 0.09 \\
            B6 & Y & N & \textbf{\underline{Y}} & N & 0.73 $\pm$ 0.02 \\
            B7 & Y & \textbf{\underline{Y}} & Y & N & 0.74 $\pm$ 0.05 \\
            B8 (GIN) & Y & Y & Y & \textbf{\underline{Y}} & 0.76 $\pm$ 0.03 \\
            \bottomrule
        \end{tabular}
    \end{adjustbox}
    \label{tab:adjustment}
    \vspace{-0.5cm}
\end{table}

Table. \ref{tab:adjustment} reveals how much each module contributes to the performance improvement. By comparing B1-B3 and B4-B8, the impact of modifying each module is evaluated depending on whether costs are used for constraint policy optimization. The differences from B1 to B2, from B4 to B5, and from B6 to B7 represent the degree of improvement by using weighted edge (Wedge) in Eq. \ref{eq:edge}. Comparing B2 to B3 and B5 to B6, the distance average pooling (DAP) in Eq. \ref{eq:dap} shows a significant improvement because it helps ego-centered control by paying attention to spatially nearby objects. Finally, B8 verifies the advantage of dense signals with auxiliary cost (Aux. Cost).

\subsection{Visualization}
\begin{figure}[!t]
    \centering
    \includegraphics[width=0.85\columnwidth]{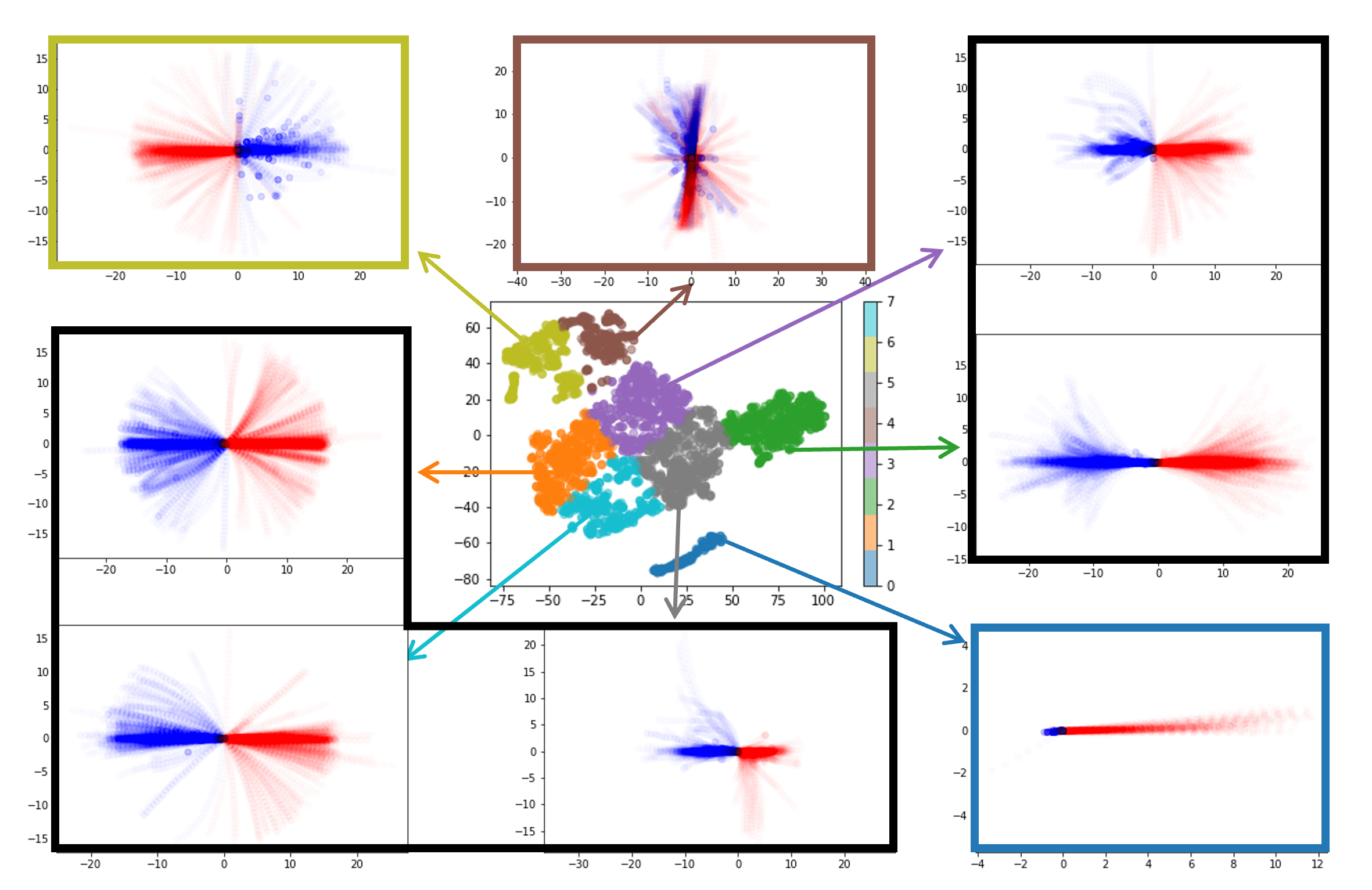}
    \caption{Visualization of social context with trajectory distributions. Each axis represents a value for x and y in the Cartesian coordinate system. The color for distinguishing the social context indicates the label for each trajectory distribution. In the trajectory distribution, the blue portion denotes the previous history, whereas the red portion represents the predicted future.}
    \label{fig:vis_social_contex}
    \vspace{-0.5cm}
\end{figure}

To illustrate the social context vectors, we cluster the distribution using t-distributed stochastic neighbor embedding (t-SNE). The trajectory distribution for each cluster is divided according to the direction and shape, as shown in \ref{fig:vis_social_contex}. Specifically, the 6th cluster (yellow) is a distribution that moves in the opposite direction to the ego-vehicle, the 4th distribution (brown) moves in the vertical direction, and the 0th distribution (blue) refers to the intention to stand still and move forward. Others (black) represent the same direction, but have slightly different shapes. This is because most of the patterns encountered are in the same direction, and the latent vectors are in similar locations.

\section{Conclusion}
We propose an approach that efficiently combines motion prediction and control modules by learning shared inputs and creating two feedback loops via a safe RL framework. To summarize the dynamic scene, the shared variable reflects the spatiotemporal locality by alternately passing through a multilayer graph operation and temporal convolution. Furthermore, we enable policies to avoid long-term risks by using cost signals obtained from prediction results as constraints. Incorporating the two modules through a safe RL framework makes predictions robust in risky situations. This significantly improves the performance of navigation problems with dense traffic, which is limited in the existing baselines. In addition, visualization of the learned latent variables shows how our method captures interactions.

In the future, we will extend our work to reflect more road context information in graph form. Another exciting direction would be to investigate the multimodal intentions of the surrounding heterogeneous agents.

\addtolength{\textheight}{-12cm}   




\bibliographystyle{IEEEtran}
\bibliography{IEEEabrv, ref}

\end{document}